\documentclass[runningheads]{llncs}
\usepackage{graphicx}
\graphicspath{{./}} 
\usepackage{comment}
\usepackage{amsmath,amssymb} 
\usepackage{color}

\usepackage{microtype}
\usepackage{acronym}
\usepackage[dvipsnames]{xcolor}
\usepackage{caption}
\usepackage{hyperref}
\usepackage{wrapfig}
\newcommand{\mat}[1]{\mathbf{#1}}

\newcommand{\R}{\mathbb{R}}
\newcommand{\h}{\mathbf{h}}
\newcommand{\s}{\mathbf{s}}

\newcommand{\myparagraph}[1]{\smallskip\noindent\textbf{#1.}}
\newcommand{\mysubparagraph}[1]{\smallskip\noindent-- \emph{#1:}}
\acrodef{STMPNN}[\textsc{VS-ST-MPNN}]{Visual Symbolic - Spatio Temporal - Message Passing Neural Network}
\acrodef{GNN}[\textsc{GNN}]{Graph Neural Network}

\usepackage[normalem]{ulem}

\begin{document}
\pagestyle{headings}
\mainmatter
\def\ECCVSubNumber{6535}  

\title{Representation Learning on Visual-Symbolic Graphs for Video Understanding} 

\titlerunning{Representation Learning on Visual-Symbolic Graphs}
%
\index{B\'ejar Haro, Benjam\'in}
\author{Effrosyni Mavroudi\orcidID{0000-0001-7552-8342} \and
Benjam\'in B\'ejar Haro\orcidID{0000-0001-9705-4483} \and
Ren\'e Vidal\orcidID{0000-0003-1838-0761}}
\authorrunning{E. Mavroudi et al.}
\institute{Mathematical Institute for Data Science, Johns Hopkins University, Baltimore, MD \\
\email{\{emavrou1,bbejar,rvidal\}@jhu.edu}}
\maketitle

\begin{abstract}
Events in natural videos typically arise from spatio-temporal
interactions between actors and objects and involve multiple co-occurring
activities and object classes. 
To capture this rich visual and semantic context, we propose using two graphs:
(1) an attributed spatio-temporal visual graph whose nodes correspond to
actors and objects and whose edges encode different types of interactions,
and (2) a symbolic graph that models semantic relationships.
We further propose a graph neural network for refining the representations of 
actors, objects and their interactions on the resulting hybrid graph. Our 
model goes beyond current approaches that assume nodes and edges are of the 
same type, operate on graphs with fixed edge weights and do not use a symbolic graph. 
In particular, our framework: 
a) has specialized attention-based message 
functions for different node and edge types; b) uses visual edge features; 
c) integrates visual evidence with label relationships; and d) performs global 
reasoning in the semantic space.
Experiments on challenging video understanding tasks, such as temporal action 
localization on the Charades dataset, show that the proposed method leads to 
state-of-the-art performance.
\end{abstract}


\section{Introduction}
\label{sec:intro}
\begin{figure}[t]
\centering
\includegraphics[trim={0 0cm 0 0},clip,scale=0.3]{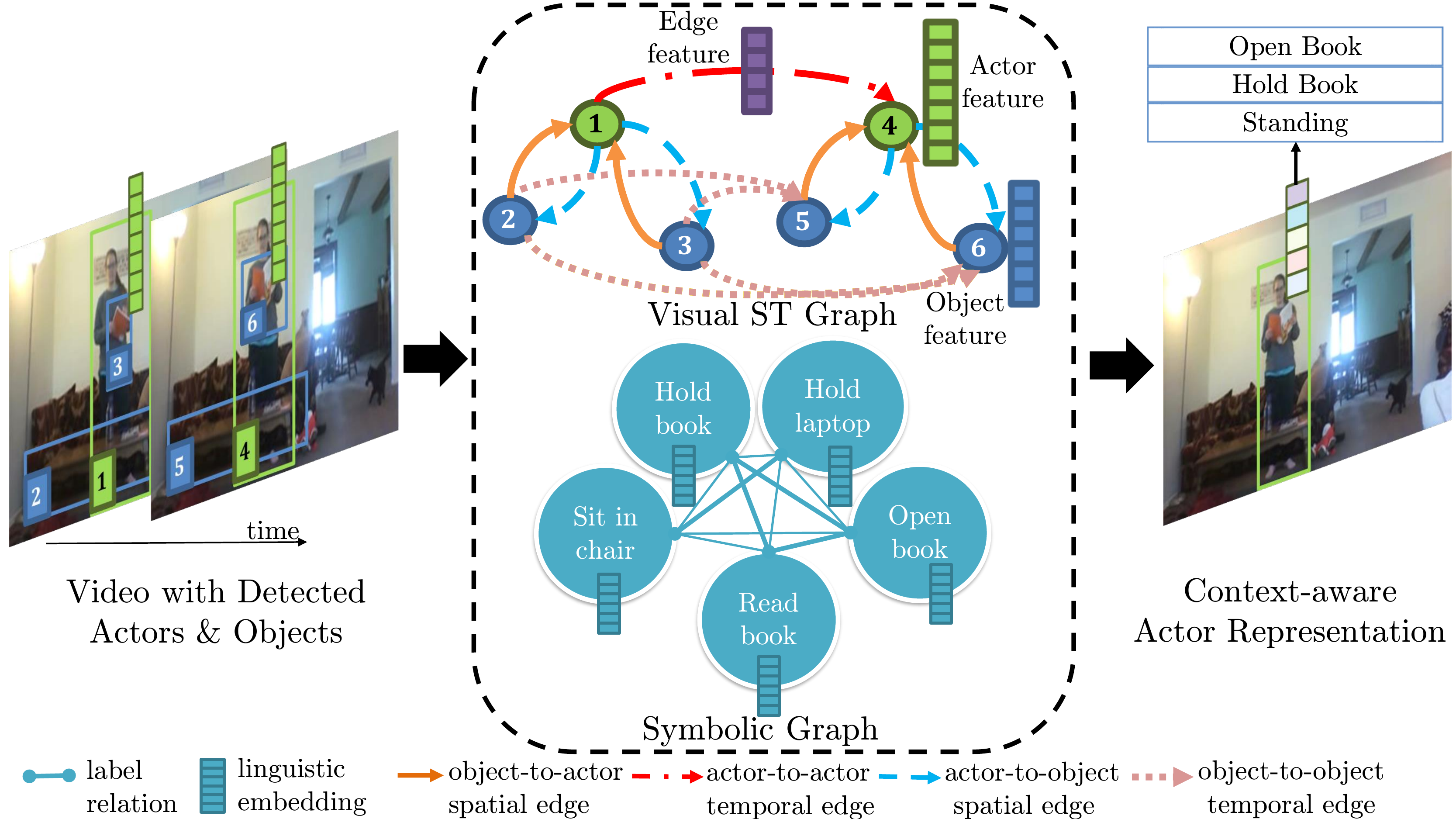}

   \caption{\textbf{Cues for video understanding:} (1) \emph{visual spatio-temporal interactions} between actors and objects and 
   (2)  commonsense 
   \emph{relationships between labels}, such 
   as co-occurrences. These cues can be encoded in a hybrid spatio-temporal 
   visual and symbolic attributed graph. In this work, we perform 
   representation learning on this hybrid graph to obtain context-aware 
   representations of detected semantic entities, such as actors and objects, that can be used to solve downstream video understanding tasks, such as multi-label action recognition.}
\label{fig:motivation}
\end{figure}
The field of video understanding has been moving towards increasing levels of 
complexity, from classifying a single action in short videos to 
detecting multiple complex activities performed by multiple actors 
interacting with objects in untrimmed videos. Therefore, there is a need to develop 
algorithms that can effectively model spatio-temporal visual and semantic 
context. One way of capturing such context is to use graph-based modeling, 
which has a rich history in computer vision.  Traditional graph-based approaches to video understanding, e.g., using probabilistic graphical models~
\cite{Koller:CVPR94,Koppula:IJRR13,Zhu:CVPR13,Wang:CVPR15}, focused mainly on modeling context at the 
level of symbols rather than visual signals or representations. However, recent 
advances have enabled \emph{representation learning on 
graph-structured data} using deep architectures called Graph Neural Networks 
(GNNs), which learn how to refine node representations by 
aggregating messages from their neighbors~\cite{Kipf:ICLR17}.

Videos can be represented as visual spatio-temporal attributed graphs (visual 
st-graphs) whose nodes correspond to regions obtained by an object detector
and whose edges capture interactions between such regions. GNNs have recently 
been designed for refining the local node/edge features, typically extracted 
by a convolutional neural network, based on the spatio-temporal context 
captured by the graph. Although representation learning on visual st-graphs 
has lead to significant advances in video understanding~
\cite{Zhang:CVPR19,Sun:ECCV18,Girdhar:CVPR19,Wang:ECCV18,Huang:BMVC19,Yuan:ICCV17,Baradel:ECCV18}, 
there are four key limitations of state-of-the-art approaches that prevent 
them from fully exploiting the rich structure of these graphs.
First, the visual st-graph is a \emph{heterogeneous} graph that has 
distinct node types (\emph{actor}, \emph{object}, etc.) and distinct edge types (\emph{object-to-actor spatial}, \emph{actor-to-actor temporal}, etc.), with each type being 
associated with a feature of potentially different dimensionality and 
semantics, as shown in the example of Fig.~\ref{fig:motivation}. However, most GNNs assume nodes/edges of the same type. Therefore, recent attempts at explicitly modeling actors and objects 
have resorted to applying separate GNNs for each node/edge type~\cite{Zhang:CVPR19,Ghosh:WACV20}.
Second, most methods operate on a graph of fixed edge weights with dense 
connectivity between detected regions. 
In practice, only a few of the edges capture meaningful interactions. 
Third, current approaches do not incorporate edge features, such as \emph{geometric relations between regions}, for updating the 
node representations. Finally, despite modeling local visual context, 
existing approaches do not reason at a global video level or exploit semantic label relationships, which have been shown to be 
beneficial in the image recognition domain~\cite{Liang:NIPS18,Chen:CVPR18}.

In this work, in an effort to address these limitations, we propose a novel 
\ac{GNN}  model, called \ac{STMPNN}, that performs representation learning on 
visual st-graphs to obtain context-aware representations of detected actors 
and objects (Fig.~\ref{fig:motivation}). Our model handles heterogeneous graphs by employing \emph{learnable 
message functions that are specialized for each edge type}.
We also adapt the visual edge weights with an \emph{attention mechanism} 
specialized for each type of interaction. For example, an actor node will 
separately attend to actor nodes at the previous frame and object 
nodes at the current frame. Furthermore, we use \emph{edge features} to refine the actor and object representations and to compute the attention coefficients that determine the connection strength between regions. Intuitively, nodes which are close to 
each other or are interacting should be strongly connected. Finally, one of 
our key contributions is incorporating an attributed \emph{symbolic 
graph} whose nodes correspond to semantic labels, such as actions, described 
by word embeddings and whose edges capture label relationships, such as co-occurrence. We fuse the information of the two graphs with 
learnable association weights between their nodes and learn global semantic interaction-aware features. Importantly, we do not require 
ground truth annotations of objects, tracks or semantic labels for each 
visual node. 

In summary, the contributions of this work are three-fold. 
First, we model contextual cues for video understanding by combining a 
symbolic graph, capturing semantic label relationships, with a visual st-graph, encoding interactions between detected actors and objects. 
Second, we introduce a novel \ac{GNN} that can perform joint representation 
learning on the heterogeneous visual-symbolic graph, in order to obtain visual and 
semantic context-aware representations of actors, objects and their interactions in a video, which can then be used to solve downstream 
recognition tasks. 
Finally, to demonstrate the effectiveness and generality of our method, we
evaluate it on tasks such as multi-label temporal activity localization, 
object affordance detection and grounded video description on three 
challenging datasets and show that it achieves state-of-the-art performance.

\section{Related work}
\label{sec:related_work}

\myparagraph{Visual context for video understanding} Context and its role in vision has been studied for a long
time~\cite{Oliva:TCS07}.
There are two major, complementary ways of utilizing context in video understanding tasks: (a) extracting global \emph{representations from whole frames} by applying convolutional neural networks to short video segments~\cite{Simonyan:NIPS14,Tran:ICCV15,Wang:ECCV16,Carreira:CVPR17,Zhou:CVPR19} followed by long-term temporal models~\cite{Lea:CVPR17,Piergiovanni:CVPR18,Zhou:CVPR19}  
and (b) extracting \emph{mid-level representations} based on semantic parts, such as body parts~\cite{Cheron:ICCV15,Mavroudi:WACV17}, latent attributes~\cite{Liu:CVPR11}, secondary regions~\cite{Gkioxari:ICCV15}, human-object interactions~\cite{Prest:TPAMI12,Zhou:CVPR15} and object-object interactions~\cite{Ma:CVPR18,Baradel:ECCV18,Zhou:CVPR19}. Our proposed method falls into the latter category, using GNNs to obtain representations of detected semantic entities based on interactions captured by visual and symbolic graphs.

\myparagraph{Graph neural networks for video understanding}
The first approach applying a deep network on a visual graph for video understanding was the Structured Inference Machine~\cite{Deng:CVPR16}, which introduced actor feature refinement with message passing, and trainable gating functions for filtering out spurious interactions, but only captured spatial relationships between actors. 
Another early approach was the S-RNN~\cite{Jain:CVPR16}, which introduced the concept of weight-sharing between nodes or edges of the same type, but did not iteratively refine node representations. 
With the advent of \ac{GNN}s, many researchers have explored modeling whole frames~\cite{Zhou:ECCV18}, tracklets~\cite{Zhang:CVPR19}, feature map columns~\cite{Sun:ECCV18,Girdhar:CVPR19,Nicolicioiu:NIPS19} or object proposals~\cite{Wang:ECCV18,Huang:BMVC19,Yuan:ICCV17} as graph nodes and using off-the-shelf GNNs, such as MPNNs~\cite{Gilmer:ICML17}, GCNs~\cite{Kipf:ICLR17} and Relation Networks~\cite{Ibrahim:ECCV18,Sun:ECCV18,Zhou:ECCV18,Baradel:ECCV18} to refine the node or edge representations, obtaining significant performance gains. However, most of these \ac{GNN}s are unable to handle edge features, directed edges and distinct node and edge types.
Therefore, applying existing GNNs to visual st-graphs requires treating every node and edge in the same way~\cite{Baradel:ECCV18,Girdhar:CVPR19}, or focusing only on one edge type~\cite{Sun:ECCV18,Zhou:ECCV18,Ma:CVPR18,Huang:BMVC19,Ibrahim:ECCV18}, or using separate GNNs for each type of interaction~\cite{Wang:ECCV18,Zhang:CVPR19,Ghosh:WACV20}, hence completely ignoring or sub-optimally handling their rich graph structure. In contrast, our proposed method can be directly applied to any st-graph and supports message passing in heterogeneous graphs. The benefit of such fine-grained modeling has already been established in fields such as computational pharmacology and relational databases~\cite{Zitnik:Bioinformatics18,Gong:CVPR19,Schlichtkrull:ESWC18}, but remains relatively unexplored in computer vision. Furthermore, similar to~\cite{Qi:ECCV18,Girdhar:CVPR19}, our method iteratively adapts the visual edge weights, but employs an attention mechanism that is specialized for different edge types and takes edge features into account.

\myparagraph{Symbolic graphs} There is a long line of work on exploiting
external knowledge encoded in label relation graphs for visual recognition
tasks. Semantic label hierarchies, such as co-occurrence,
have been leveraged for
improving object recognition~\cite{Marszalek:ECCV08,Marszalek:CVPR07,Choi:CVPR10,Deng:ECCV14}, multi-label zero-shot learning~\cite{Lee:CVPR18} and other image-based visual tasks~\cite{Li:ICCV17,Ramanathan:CVPR15}. Much fewer papers utilize knowledge graphs for video understanding~\cite{Assari:CVPR14,Jiang:TPAMI18,Junior:TPAMI19}, possibly due to the limited number of semantic classes in traditional video datasets. However, most of these methods directly perform inference on the symbolic graph. For example, the SINN~\cite{Junior:TPAMI19} performs graph-based inference in a hierarchical label space for action recognition. 
Rather, we aim to use the semantics of labels to integrate prior knowledge about the inter-class relationships and facilitate the computation of semantic context-aware region features. In a similar vein, Liang et al.~\cite{Liang:NIPS18} enhance feature maps extracted from images by using a symbolic graph, while~\cite{Li:NIPS18,Chen:CVPR19} use a latent interaction graph. In contrast, we seek to improve the representation of visual st-graph nodes rather than enhance convolutional features on a regular grid. Fusing information from multiple graphs using GNNs is an exciting new line of research~\cite{Bajaj:ICCV19,Li:ICCV19,Xiong:ICCV19,Yu:NIPS19,Teney:CVPR17}. Similar to our approach, Chen et al.~\cite{Chen:CVPR18} combine a
visual graph instantiated on objects with a symbolic graph and perform graph representation learning, while~\cite{Jiang:NIPS18} enforce the scalar edge weights between visual regions to be consistent with the edges of the symbolic graph. However, they operate on simple spatial graphs and assume access to semantic labels of regions during training.


\section{Method}
\label{sec:method}
\begin{figure}[ht]
\begin{center}
 \includegraphics[trim={0 0.6cm 0 0},clip,scale=0.36]{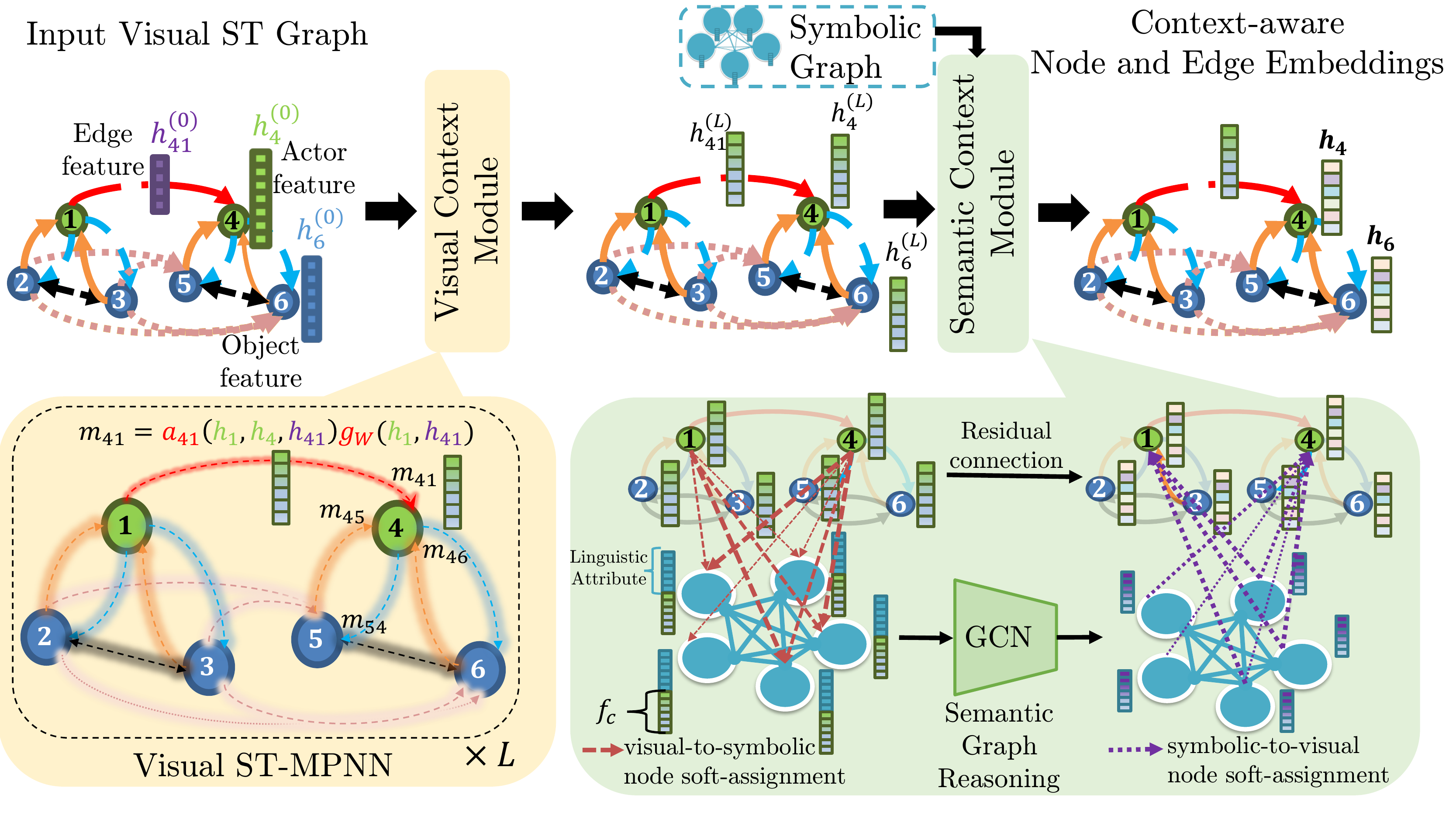}
\end{center}
   \caption{\textbf{Overview of our \ac{STMPNN} model that performs representation learning on a hybrid visual-symbolic graph}. Given an input video that is represented as a visual st-graph, with nodes corresponding to detected actors and objects and edges capturing latent interactions, our framework has two modules that integrate context in the local representations of its nodes and edges: (a) a Visual Context Module (Sec.~\ref{subsec:visual_context}) that performs $L$ rounds of node and edge updates on the visual graph, with specialized neighborhood aggregation functions that depend on the type of an edge and (b) a Semantic Context Module (Sec.~\ref{subsec:semantic_context}) that integrates visual evidence with semantic knowledge encoded in an external symbolic graph and learns global semantic interaction-aware features.
   }
\label{fig:overview}
\end{figure}
In this section, we describe the overall architecture of our proposed \ac{STMPNN} model, shown in Fig.~\ref{fig:overview}. Our goal is to refine the features of detected actors, objects and their interactions based on the contextual information captured in two graphs: a visual st-graph
and a symbolic graph. The refinement is performed by a novel GNN, which is designed to exploit the rich structure of the visual st-graph by utilizing edge features and learning specialized attention-based neighborhood aggregation functions for different node and edge types.  In addition, our model enables the fusion with the symbolic graph, by incorporating graph convolutions, to learn semantic relation-aware features, and soft-assignment weights, to connect visual and symbolic graph nodes without requiring access to ground-truth semantic labels of regions during training. Our model can be trained jointly with recognition networks to solve downstream video understanding tasks.

\subsection{Visual Context Module}
\label{subsec:visual_context}
\myparagraph{Visual st-graph} Our 
input is a sequence of $T$ frames with detected actor and object regions. Let $G^v = (V^v, 
E^v)$ be a spatio-temporal attributed directed 
graph, called the 
\emph{visual st-graph}, where $V^v$ is a finite set 
of vertices and $E^v 
\subseteq V^v \times V^v$ is a set of edges. Nodes correspond to actor and object detections, while edges model latent interactions. There are $M$ actors and $N$ objects per frame. Fig.~\ref{fig:overview} illustrates a toy example with $M=1$, $N=2$ and $T=2$.
The graph is both node- and edge-typed
with $\mathcal{N}$ node types and $\mathcal{E}$ edge types, i.e., each node (edge) is associated with a single node (edge) type. Specifically, the node types are
\emph{actor} and \emph{object} ($\mathcal{N}=2$) and the edge types are object-to-actor spatial (\emph{obj-act-s}), actor-to-object spatial (\emph{act-obj-s}), object-to-object spatial (\emph{obj-obj-s}), actor-to-actor temporal (\emph{act-act-t}) and object-to-object temporal (\emph{obj-obj-t})  ($\mathcal{E}=5$). The allowed spatio-temporal connections between nodes of the visual st-graph ($E^v$) are specified a priori and encode the family of spatio-temporal interactions captured by the model. For instance, we can constrain temporal edges to connect a node 
at frame $t$ with another node of the same type at time $t-1$.
Each node and edge is described by an initial attribute, whose dimensionality may vary depending on the node/edge type. An actor/object appearance feature can be used as the initial attribute of node $i$ ($\h^{(0)}
_i$), while the relative spatial location of regions $i$ and $j$ can be used as the initial attribute of the edge from $j$ to $i$ ($\h^{(0)}
_{ij}$).

\myparagraph{Visual ST-MPNN}
Given the input visual st-graph $G^v$ with initial node and edge attributes/features, $\{\h_{i}
^{(0)}\}_{i \in V^v}$ and $\{\h_{ij}^{(0)}\}_{(i, j) \in E^v}$,
respectively, we introduce novel GNN propagation rules to perform representation learning on the visual st-graph with the goal of refining local node and edge features using spatio-temporal contextual cues.
At each iteration our model: (1) refines the scalar visual edge weights using attention coefficients;
(2) computes a message along each edge that depends on the edge type, the attention-based  scalar edge weight, the features of the connected nodes and the edge feature; (3) updates the feature of every node by aggregating messages from incoming edges; and (4) updates the feature of every edge by using the message that was computed alongside it. Next, we describe each one of these steps in more detail.

\mysubparagraph{Attention mechanism} At each iteration $l$ of the 
MPNN, we first refine the strength of region connections by computing \emph{attention 
coefficients}, $a_{ij}$, that capture the relevance of node $j$ (message sender) for the update 
of node $i$ (message receiver). In contrast to GAT~\cite{Velickovic:ICLR18}, our model learns an attention mechanism specialized for each type of interaction and it utilizes edge features for its computation.
The attention coefficients for the $l$-th iteration are computed as follows:
\begin{equation}
a_{ij}^{(l)} = \exp{\left(\gamma^{(l)}_{ij}\right)} / \left( \sum_{k \in N^v_{\epsilon_{ij}}(i)} \exp{\left(\gamma^{(l)}_{ik}\right)} \right) ,
\end{equation}
\begin{equation}
\gamma^{(l)}_{ij} = \rho \left(  \left({\mathbf{v}_a ^{\epsilon_{ij}} }\right) ^ { T } \left[ W_r^{\nu_i} \h _ { i }^{(l-1)} ; W_s^{\nu_j}  \h _ { j }^{(l-1)} ; \beta W_{e}^{\epsilon_{ij}} \h _ { ij }^{(l-1)} \right] \right) .
\label{eq:attention}
\end{equation}
Here, $\epsilon_{ij}$ is the type of the 
edge from node $j$ to node $i$, $N^v_{\epsilon_{ij}}(i)$ is the set of visual nodes connected with node $i$ via an incoming edge of type $\epsilon_{ij}$,
$\h_{i}^{(l-1)}$ is the feature of the $i$-th node at the previous iteration,
$\h_{ ij }^{(l-1)}$ is the feature of the edge from $j$ to $i$ at the previous iteration, $\nu_i$ is the type of node $i$ and $\rho$ is a non-linearity, such as Leaky-ReLU~\cite{He:ICCV15}. The parameter $\beta \in \{0,1\}$ denotes whether edge features will be used for computing the attention coefficients ($\beta=1$) or not $(\beta=0)$.
$W_{r}^{\nu_i}$, $W_{s}^{\nu_j}$ and  $W_{e}^{\epsilon_{ij}}$ are 
learnable projection weights and are shared between nodes (edges) of the same type. All projection matrices linearly transform the current node 
(edge) feature to a refined feature of fixed dimensionality $d_l$. $
\mathbf{v}_a ^{\epsilon_{ij}}$ is a learnable attention vector.
For improved 
readability we have dropped the layer index ($l$) from the attention and 
projection weights.

\mysubparagraph{Message computation} After computing the attention coefficients, we compute a message along each edge. The message from node $j$ to node $i$ is:
\begin{equation}
\mathbf{m}^{(l)}_{ij} = a_{ij}^{(l)} \left(\lambda_v W_s^{\nu_j} \h _ { j }^{(l-1)} + \lambda_e  W_{e}^{\epsilon_{ij}}  \h _ { ij }^{(l-1)} \right),
\label{eq:message_across_edge}
\end{equation}
where the parameters $\lambda_e,\lambda_v\in\{0,1\}$ denote whether the edge feature and the sender node feature will be used for the message computation, respectively. 

\mysubparagraph{Node and edge update} Following the message computation, the 
node feature is updated using an aggregation of incoming messages from different edge types and a residual 
connection, while the 
edge feature is set to be equal to the message:
\begin{equation}
\h _ { i }^{(l)} = \h _ { i }^{(l-1)} + \sigma\left(\sum_{j \in N^v(i)} \mathbf{m}^{(l)}_{ij}\right),
\h _ { ij }^{(l)} = \mathbf{m}^{(l)}_{ij},
\end{equation}
where $N^v(i)$ is the set of visual nodes that are connected with node $i$, $\sigma(\cdot)$ is a non-linearity, such as ReLU. After $L$ layers of the spatio-temporal MPNN (or equivalently $L$ rounds of node and edge 
updates), we obtain refined, visual context-aware node and edge 
features: $\h _ { i }^{(L)}\in \R^{d_L}$ and $\h _ { ij }
^{(L)} \in \R^{d_L}$.

\subsection{Semantic Context Module}
\label{subsec:semantic_context}

\myparagraph{Symbolic graph} 
Let  $G^s = (V^s, E^s)$, be the input \emph{symbolic} graph, where $V^s$ and 
$E^s$ denote the symbol set and edge set, respectively. The nodes of this 
graph correspond to semantic labels, such as action labels or object labels. 
Each symbolic node $c$ is associated with a semantic attribute, such as the 
linguistic embedding of the label ($\s_c \in \R^K$). Edges in the symbolic graph are associated with scalar weights, which encode label relationships, such as co-occurrence. 
These edge weights are summarized in the fixed 
adjacency matrix $L^s \in \mathbb{R}^{|V^s| \times |V^s|}$.

\mysubparagraph{Integration of visual evidence with the symbolic graph}
As a first step, we update the attributes of the symbolic graph using visual evidence, 
i.e., the visual context-aware representations of the nodes of the visual st-graph.
To achieve this, without requiring access to the ground-truth semantic labels of regions, we learn 
associations between the nodes of the visual st-graph and those of the symbolic graph.
These associations are the edges of the bipartite graph $G^{vs} = (V^{vs}, E^{vs})$, with $ V^{vs} = V^v \cup V^s$ and $E^{vs} \subseteq V^v \times V^s$. Although latent, we can specify a priori the allowed visual-to-symbolic node connections (edges). For example, when symbolic nodes correspond to action classes, we can remove edges between object and symbolic nodes.
The learnable association weight $\phi_{c,i}^{vs}$ represents the confidence of assigning the feature from visual node $i$ to the symbolic node $c$:
\begin{equation}
    \phi^{vs}_{c, i} = \frac{\exp\left(\left(\mat{w}_c^{vs}\right)^T \h_i^{(L)}\right)}{\sum_{c^\prime \in N^{vs}(i)} \exp\left(\left(\mat{w}_{c^\prime}^{vs}\right)^T \h_{i}^{(L)}\right)},
\end{equation}
where $\mat{w}_c^{vs} \in \R^{d_L}$ is a 
trainable weight vector and $N^{vs}(i)$ is the neighborhood of visual node $i$ on the bipartite graph $G^{vs}$.
After computing the voting weights, each symbolic node is associated with a weighted sum of projected visual node features:
$\mathbf{f}_c = \sigma(\sum_i  \phi^{vs}_{c, i} W_p^{vs} \h_i^{(L)}$),
where $W_p^{vs} \in \R^{D_s \times d_L}$ is a learnable projection weight matrix.  
The new representation of each node $c$ is computed as the 
concatenation of the linguistic embedding and the visual feature: $\s_c^{(0)} = \left[\s_c ; 
\mathbf{f}_c \right] \in \R^{K + D_s}$.

\mysubparagraph{Semantic graph convolutions} We obtain semantic relation-aware features by applying a vanilla GCN~\cite{Kipf:ICLR17} to the nodes of the symbolic graph. More specifically, by iteratively applying the propagation rule $S^{(r+1)} = \mathrm{GCN} (S^{(r)}, L^s)$, where $S^{(r)}$ denotes the matrix of symbolic node embeddings at iteration $r=1,\dots, R$, the GCN yields evolved symbolic node features $S^{(R)} \in \mathbb{R}^{|V^s| \times D_s}$.

\mysubparagraph{Update of visual st-graph} 
The evolved symbolic node representations obtained after $R$ iterations of 
graph convolutions on the symbolic graph can be mapped back to the visual st-graph, so that the representation of the visual nodes can be enriched by global semantic context. To achieve this we compute mapping weights (attention coefficients) from symbolic nodes to visual nodes:
\begin{equation}
\phi_{i,c}^{sv} = \frac{\exp{\left( \left(\mathbf{v}_a^{sv}\right)^T \left[\s_c^{(R)};\h_i^{(L)}\right]\right)}}{\sum_{c^\prime \in N^{vs}(i)} \exp{\left( \left(\mathbf{v}_a^{sv}\right)^T \left[\s_{c^\prime}^{(R)};\h_i^{(L)}\right]\right)}} ,
\end{equation}
where $\mathbf{v}_a^{sv} \in \R^{d_L + D_s}$ is a learnable attention vector.
The final visual node feature representation is then obtained using a residual connection: $	\h_i = \h_i^{(L)} + \sigma\left(\sum_{c^\prime \in V^s} \phi^{sv}_{i,c^\prime} W_p^{sv} \s_{c^\prime}^{(R)} \right)$. These context-aware representations can be fed to recognition networks to solve downstream video understanding tasks. 

\section{Experiments}
\label{sec:experiments}
To demonstrate the effectiveness and generality of our method, we conduct experiments on three challenging video understanding tasks that require reasoning about interactions between semantic entities and relationships between classes: a) sub-activity and object affordance classification (Sec.~\ref{subsec:cad120}), b) multi-label temporal action localization (Sec.~\ref{subsec:charades}) and c) grounded video description (Sec.~\ref{subsec:activitynet}).
\begin{table}[t]
\caption{\textbf{Results on CAD-120~\cite{Koppula:IJRR13}} for sub-activity and object
affordance detection, measured via F1-score. Our results are averaged over five random runs, with the standard deviation reported in parentheses.}
\label{tab:cad120_sota}
\centering
\scalebox{1}{
\begin{tabular}{|@{\,}c@{\,}|@{\,}c@{\,}|@{\,}c@{\,}|}
\hline
Method & \multicolumn{2}{c|}{Detection F1-score (\%)} \\
\hline
  & Sub-activity & Object affordance \\
\hline\hline
ATCRF~\cite{Koppula:IJRR13} & 80.4 & 81.5 \\
S-RNN~\cite{Jain:CVPR16} & 83.2 & 88.7 \\
S-RNN~\cite{Jain:CVPR16} (multitask) & 82.4 & \textbf{91.1} \\
GPNN~\cite{Qi:ECCV18} & 88.9 & 88.8 \\
STGCN~\cite{Ghosh:WACV20} & 88.5 & -\\
\hline
\ac{STMPNN} (ours) & \textbf{90.4} ($\pm 0.8$) & \emph{89.2} ($\pm 0.3$)\\
only visual graph (ours) & 89.6 ($\pm 1.1$) & 88.6 ($\pm 0.6$)\\
\hline
\end{tabular}
}

\end{table}

\subsection{Experiments on CAD-120}
\label{subsec:cad120}
\myparagraph{CAD-120} This dataset provides 120 RGB-D videos, with each video showing a daily activity comprised of a sequence of sub-activities (e.g., \emph{moving}, \emph{drinking}) and object affordances (e.g., \emph{reachable}, \emph{drinkable})~\cite{Koppula:IJRR13}. Given temporal segments, the task is to classify each actor in each segment into one of $10$ sub-activity classes and each object into one of $12$ affordance classes. Evaluation is performed with 4-fold, leave-one-subject-out, cross-validation using F1-scores averaged over all classes as an evaluation metric.  With a visual st-graph provided by the dataset~\cite{Koppula:IJRR13} (including hand-crafted features of actors and objects and geometric relations), it is a particularly good test-bed for comparing different GNNs.

\myparagraph{Implementation details}
The visual st-graph provided with the dataset is instantiated on the actors and objects of each temporal segment of the input video and contains $5$ edge types: \emph{obj-obj-s}, \emph{obj-act-s}, \emph{act-obj-s}, \emph{act-act-t} and \emph{obj-obj-t}. We construct a symbolic graph that has nodes corresponding to the $10$ sub-activity and $12$ affordance classes, with edge weights capturing per-frame class co-occurrences in training data. The attribute of each symbolic node is obtained by using off-the-shelf word2vec~\cite{Mikolov:NIPS13} class embeddings of size $K=300$. Actor (object) nodes are connected to sub-activity (affordance) symbolic nodes.
The following hyperparameters are used in the our model: $L=4$, $R=1$, $\lambda_v=1,\lambda_e=1$ and $\beta=1$. All messages are of size $256$.
We use the sum of cross-entropy losses per node to jointly train our model and the sub-activity and affordance classifiers applied at each node of the st-graph. We train for $100$ epochs with a batch size of $5$ sequences and use the Adam learning rate scheduler with an initial learning rate of $0.001$. Dropout with a rate of $0.5$ is applied to all fully connected layers.

\myparagraph{Comparison with the state of the art}
Table~\ref{tab:cad120_sota} compares the sub-activity and affordance detection performance of our method with prior work. Our method obtains state-of-the-art results for sub-activity detection, with an average performance of $\mathbf{90.4}\%$ and a best of $\mathbf{91.3}\%$, and the second best result on affordance detection ($89.2\%$) - being only second to the S-RNN (multi-task)~\cite{Jain:CVPR16}. The S-RNN was trained on the joint task of detection and anticipation and we outperform it by $8\%$ in the sub-activity classification task. Even without using the symbolic graph, our method improves upon recent GNNs, which were applied on the same attributed visual st-graph, validating our novel layer propagation rules.
\begin{figure}[t]
   \centering\includegraphics[scale=0.28]{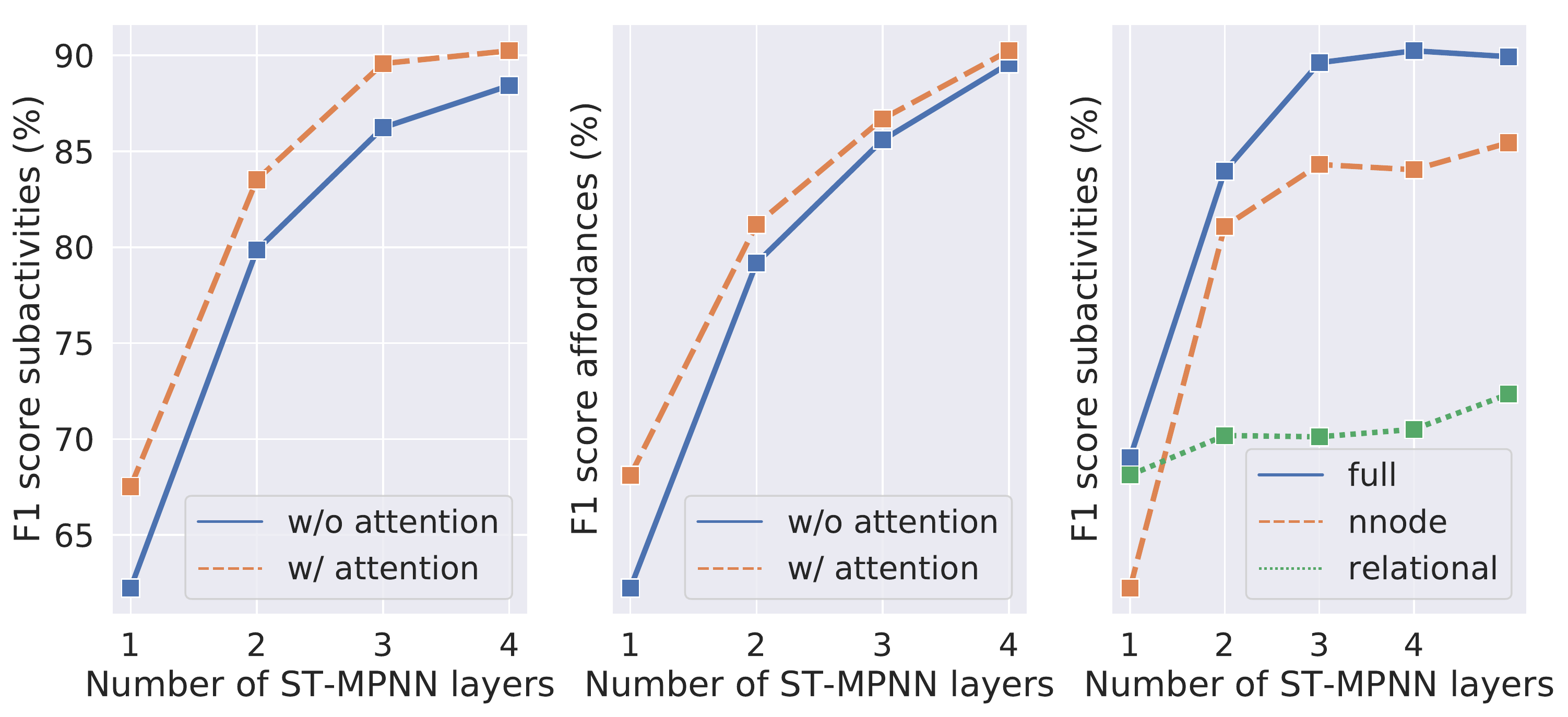}
   \caption{\textbf{Effect of attention mechanism and node update type on CAD-120 detection performance}. Using an attention mechanism  outperforms using a fixed visual adjacency matrix. Updating nodes based on both neighboring node and incoming edge attributes (\emph{full}) is superior to updating them using just the nodes (\emph{nnode}) or edges (\emph{relational}).}
\label{fig:cad120_attention_ablation}
\end{figure}
\textbf{Ablation analysis.} In Fig.~\ref{fig:cad120_attention_ablation}, we show the effect of attention, edge features and number of visual node updates on the recognition performance. First, we compare the performance of a model using a fixed binary adjacency matrix with that of a model using our attention mechanism. Clearly, attention benefits performance in both tasks. Second, we conclude that using the attributes of both the neighboring nodes and adjacent edges is better than using only those of the neighboring nodes, validating the usefulness of edge features. Finally, increasing the number of ST-MPNN layers improves performance, which saturates after 4-5 layers. 

\subsection{Experiments on Charades}
\label{subsec:charades}
\begin{table}
\caption{\textbf{Multi-label temporal action localization results on Charades~\cite{Sigurdsson:ECCV16}}. Performance is measured via per-frame mAP. 
R: RGB, F: optical flow. Our method yields a relative improvement of $6\%$ over the state-of-the-art method by using \textbf{only raw RGB frames}.}
\centering
\scalebox{1.0}{
\begin{tabular}{|@{\,}c@{\,}|@{\,}c@{\,}|@{\,}c@{\,}|@{\,}c@{\,}|}
\hline
Method & Feat & Input & mAP (\%) \\
\hline
Predictive-corrective~\cite{Dave:CVPR17} & VGG & R &  8.9\\
Two-stream~\cite{Sigurdsson:CVPR17} & VGG & R+F &  8.94\\
Two-stream + LSTM~\cite{Sigurdsson:CVPR17} & VGG & R+F &  9.6\\
R-C3D~\cite{Xu:ICCV17} & VGG & R+F & 12.7 \\
ATF~\cite{Sigurdsson:CVPR17} & VGG & R+F & 12.8 \\
\hline
RGB I3D~\cite{Piergiovanni:CVPR18}& I3D & R & 15.63 \\
I3D~\cite{Piergiovanni:CVPR18}  & I3D & R+F & 17.22 \\
I3D + LSTM~\cite{Piergiovanni:CVPR18} & I3D & R+F & 18.12 \\
RGB I3D + super-events~\cite{Piergiovanni:CVPR18} & I3D & R & 18.64 \\
I3D + super-events~\cite{Piergiovanni:CVPR18}  & I3D & R+F & 19.41 \\
STGCN~\cite{Ghosh:WACV20}  & I3D & R+F & 19.09 \\
I3D + biGRU  & I3D & R & 21.7 \\
I3D + 3TGMs + super-events~\cite{Piergiovanni:ICML19} & I3D & R+F & 22.3 \\
\hline
I3D + biGRU + VS-ST-MPNN (Ours)  & I3D & \textbf{R} & \textbf{23.7}($\pm 0.2$)\\ 
\hline
\end{tabular}
}
\label{tab:charades_sota}
\end{table}
\textbf{Charades.} Charades~\cite{Sigurdsson:ECCV16} is a large dataset with $9848$ RGB videos and temporal annotations for $157$ action classes, many of which involve human-object interactions. Each video contains an average of $6.8$ activity instances, many of which are co-occurring. Following~\cite{Sigurdsson:ECCV16}, multi-label action temporal localization performance is measured in terms of mean Average Precision (mAP), evaluating per-frame predictions for $25$ equidistant frames in each one of the 1.8k validation videos.

\myparagraph{Implementation details}
To tackle the challenging problem of multi-label temporal action localization, we perform a late fusion of a global model, operating on whole frames, and a local model, operating on actors and objects. The global model is an I3D RGB model~\cite{Carreira:CVPR17} fine-tuned on Charades~\cite{Piergiovanni:CVPR18}, combined with a two-layer biGRU of size 256, similar to existing baselines on this dataset. The proposed \ac{STMPNN} is used as the local model.
To build the visual st-graph we detect actors and objects using a Faster-RCNN~\cite{He:TPAMI18} trained on the MS-COCO~\cite{Lin:ECCV14} dataset. We rank detections based on their score and we keep the top-2 person detections and top-10 object detections per frame. 
We pool features from the \texttt{Mixed\_4f} 3D feature map of the I3D for each detected region using RoIAlign~\cite{He:TPAMI18} and max-pooling in space. This yields an attribute of size $832$ for the actor/object regions for the frames of the original video sampled at $1.5$ FPS.
We use $3$ types of visual edges: \emph{obj-act-s}, \emph{act-obj-s} and \emph{act-act-t} and describe each edge with the relative position of the connected regions. Our symbolic graph has nodes corresponding to the $157$ action classes and edge weights corresponding to per-frame label co-occurrences in training data. Only actor nodes are connected to symbolic nodes. Obtaining a linguistic attribute for each symbolic node is not trivial, since action names often contain multiple words. To circumvent that, each action class is separated into a verb and an object and the average of their word2vec~\cite{Mikolov:NIPS13} embeddings ($K=300$) is used as the initial node attribute. 
The hyperparameters are:  $L=3$, $d_L=512$, $R=1$, $D_s=256$, $\lambda_v=1,\lambda_e=1$ and $\beta=1$. For performing per-frame multi-label action classification, we average the learned actor node and edge representations at each frame, we input them to a two-layer biGRU of size 512, and we feed the resulting hidden states to binary action classifiers for per-frame multi-label action classification. We jointly train the \ac{STMPNN} and biGRU for $40$ epochs with a binary cross-entropy loss applied per frame, using a batch size of $16$ sequences. We also apply dropout with a rate of $0.5$ on all fully connected layers and use the Adam scheduler, with an initial learning rate of $1e^{-4}$.

\myparagraph{Comparison with prior work}
As shown in Table~\ref{tab:charades_sota},
our framework outperforms all other methods on temporal action localization, with a mAP of $\mathbf{23.7}\%$ (averaged across 3 random runs) by using only raw RGB frames. It yields a relative improvement of $24\%$ over the alternative graph-based approach~\cite{Ghosh:WACV20}, which uses both RGB and optical flow inputs, as well as additional actor embeddings trained at the ImSitu dataset~\cite{Yatskar:CVPR16}.

\begin{figure}[ht]
\begin{minipage}{0.45\textwidth}
   \centering\includegraphics[scale=0.27,trim={0cm 0cm 0cm 0cm},clip]{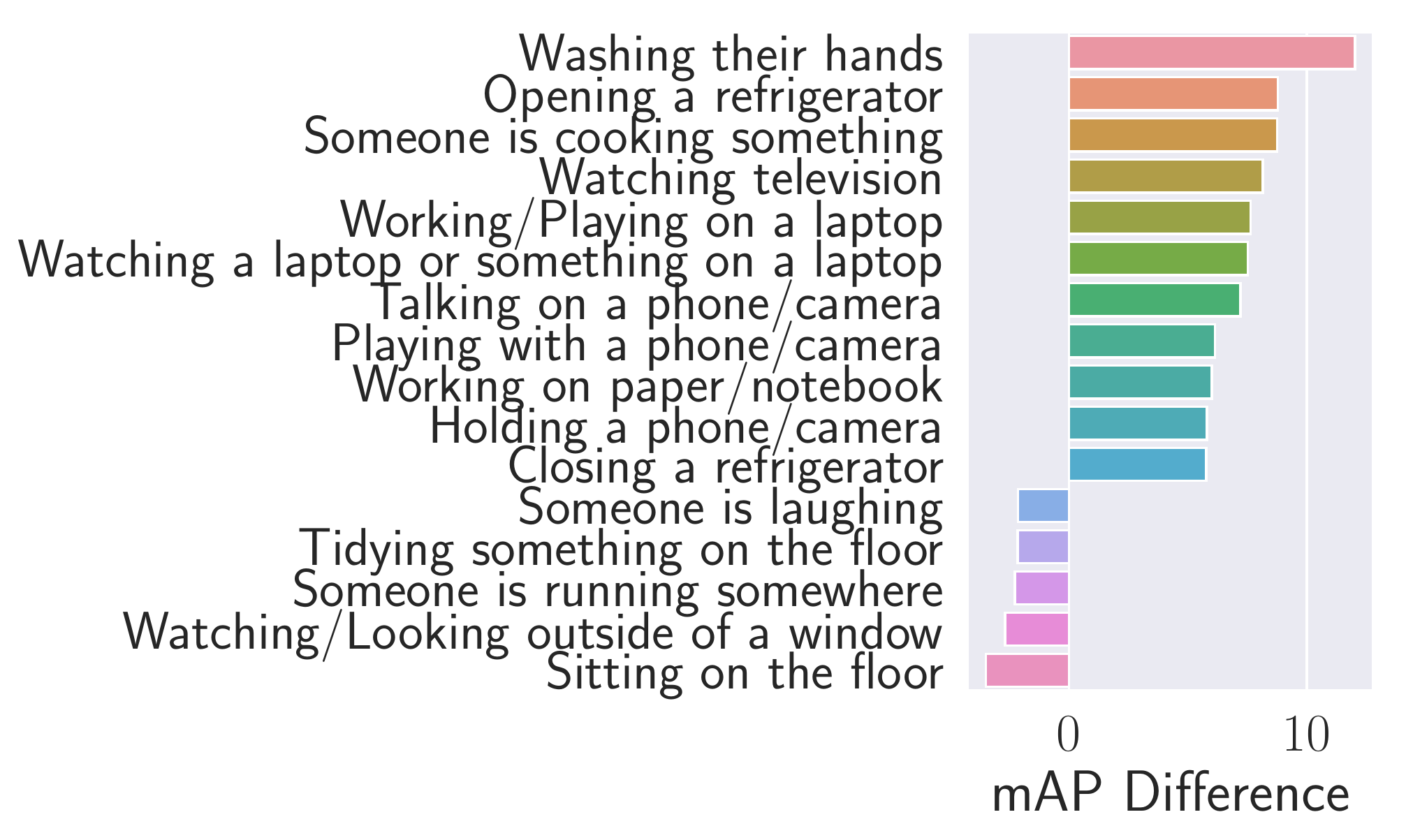}
\end{minipage}
\begin{minipage}{0.45\textwidth}
\centering\includegraphics[scale=0.32]{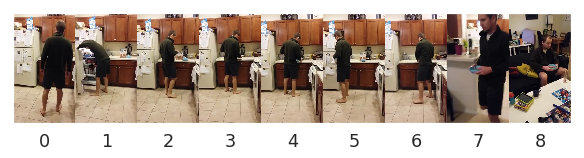}
\centering\includegraphics[scale=0.23]{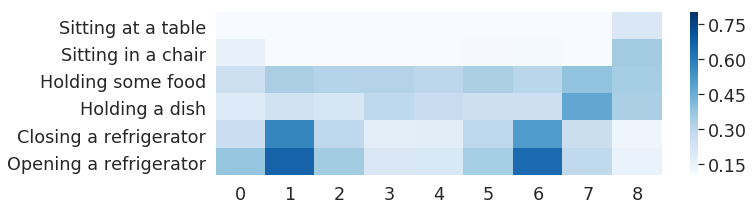}
\end{minipage}
   \caption{\textbf{Qualitative results on Charades}. (\emph{left}) The classes with the highest positive and negative performance difference after adding \emph{object-to-actor spatial} messages. Incorporating spatial structure benefits actions that involve interactions with distant objects, such as \emph{watching television} or \emph{cooking}. (\emph{right}) Action predictions of our model (ID: 3) for 9 frames of a sample Charades video.
}
\label{fig:charades_diff_per_class}
\end{figure}

\begin{figure}[ht]
\centering\includegraphics[scale=0.23]{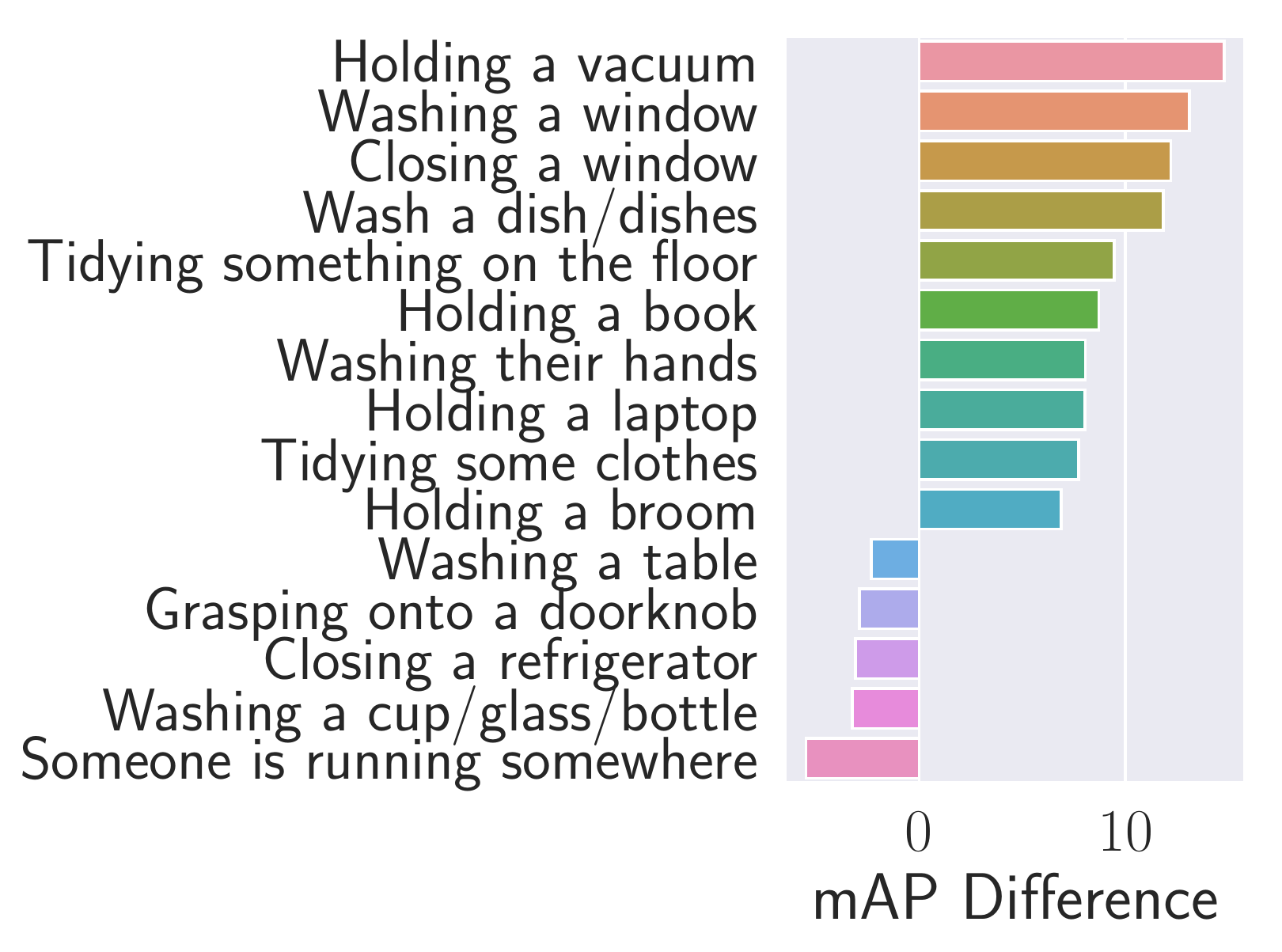}
\centering\includegraphics[scale=0.09]{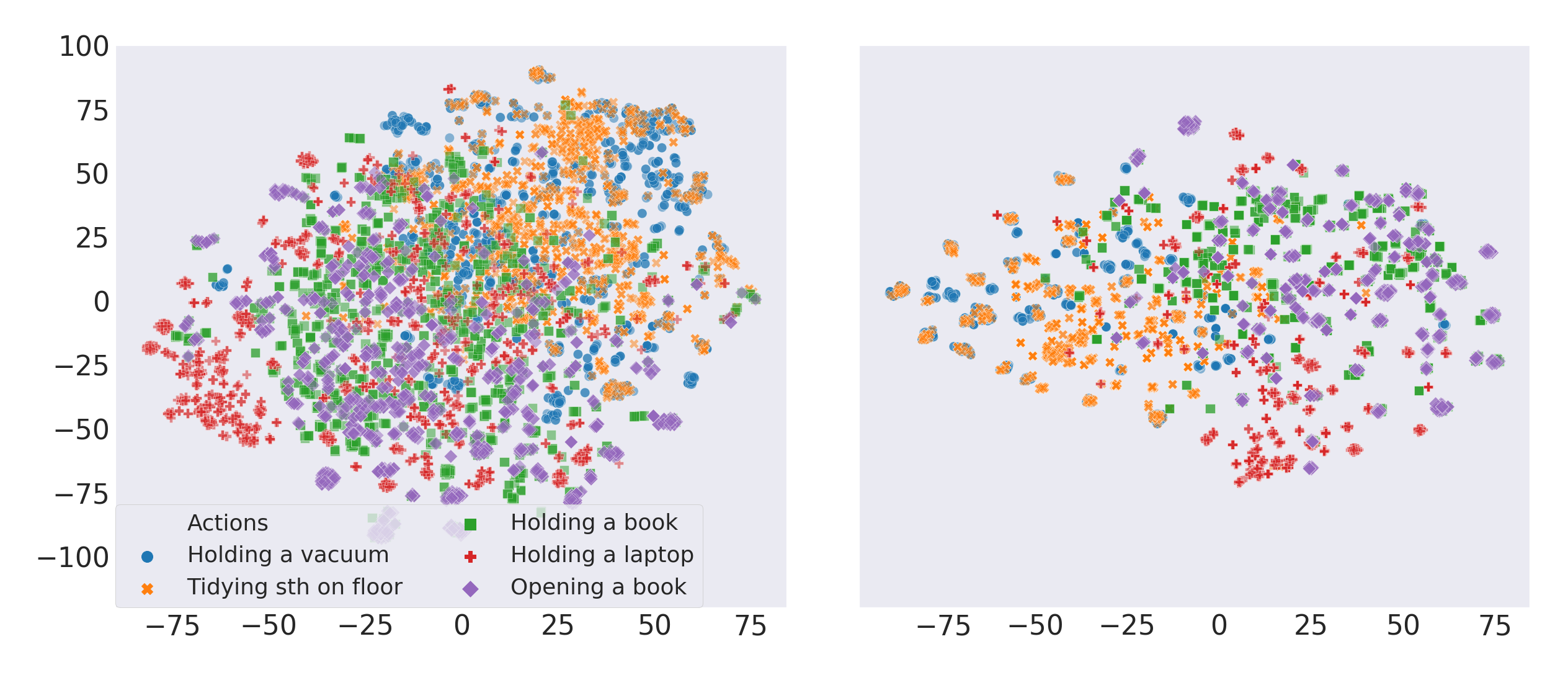}
 \caption{\textbf{Qualitative evaluation of the Semantic Context Module (SCM)}. (\emph{left}) Classes with the highest positive and negative performance difference when adding the semantic module. (\emph{right}) t-SNE visualization of actor node embeddings from Charades validation set obtained before and after adding the SCM. We show 1121 random samples per class for $5$ selected action classes. (\emph{Best viewed zoomed in and in color}.)}
\label{fig:tsne}
\end{figure}

\begin{table}[ht]
\caption{\textbf{Ablation analysis on Charades~\cite{Sigurdsson:ECCV16}}. \emph{Visual}: Visual Context Module. \emph{Semantic}: Semantic Context Module. \emph{Long Term}: long-term temporal modeling.}
\centering
\scalebox{0.8}{
\begin{tabular}{|@{\,}c@{\,}|@{\,}c@{\,}@{\,}c@{\,}@{\,}c@{\,}|@{\,}c@{\,}|@{\,}c@{\,}|}
\hline
ID & Visual & Semantic & Long Term & mAP (\%) & mAP (\%) \\
\hline
 &  & & & & + Global model \\
 \hline
1 & \checkmark & \checkmark  &  \checkmark & \textbf{18.6} & \textbf{23.4} \\
2 & - & -  &  \checkmark & 15.2 & 22.2  \\
3 & \checkmark & \checkmark  & - & 15.3 & 22.0 \\
4 & \checkmark & - & - & 13.7 & 21.8 \\
5 & - & \checkmark  & - & 11.7 & 21.8 \\
6 & - & - & - & 10.7 & 20.9 \\

\hline
\end{tabular}
}
\label{tab:charades_ablation}
\end{table}
\noindent\textbf{Impact of each graph.} In Table~\ref{tab:charades_ablation}, we report the baseline result  ($10.7\%$) obtained by classifying activities based on local actor features (ID: 6). Refining these features by using our Visual Context Module improves performance by $3\%$. As shown quantitatively in the supplementary material,  both our specialized attention mechanism and the usage of edge features improve the performance, outperforming a vanilla GNN. Representation learning on the hybrid graph yields a significant absolute improvement of \textbf{$\mathbf{5\%}$ over the baseline}. Additionally, modeling long-term temporal context and global context leads to the final state-of-the-art performance, indicating that the representations learned by our model are complementary to holistic scene cues and temporal dynamics.\\
\textbf{Per-class improvement analysis.}
To gain a better understanding of the benefits of representation learning on the visual graph,  we highlight in Fig.~\ref{fig:charades_diff_per_class} the activity classes with the highest positive and negative difference in performance when adding \emph{obj-to-act-s} messages. By harnessing visual human-object interaction cues, our model is able to better recognize actions such as \emph{Watching television}. \\
\textbf{Impact of semantic context module.} Comparing the models with IDs $3$ and $4$ in Table~\ref{tab:charades_ablation}, we observe that adding the semantic context module improves mAP by $2\%$. Notably, updating the visual nodes by attending over the initial symbolic node features (linguistic) instead of the evolved features did not improve performance in our experiments, showing the importance of semantic graph convolutions. The semantic module seems to particularly help with rare classes, such as \emph{Holding a vacuum}, which has only $213$ training examples ($3\%$ of available annotated segments), and classes with strong co-occurrences (Fig.~\ref{fig:tsne}). The t-SNE visualization shows that, although the visual context-aware actor embeddings are already capturing
meaningful label relationships (e.g., \emph{open} and \emph{hold book}), the integration of semantic relationships via the symbolic graph results in more tightly clustered embeddings and well-defined groups, facilitating action recognition.
\textbf{Model complexity.} Since our visual st-graph is designed to capture only local spatio-temporal interactions, we can compute messages in parallel and process the entire Charades validation set (around 2K videos at 1.5FPS) in
$2$ minutes on a single Titan XP GPU, given initial features pooled from actor/object regions. 
\subsection{Experiments on ActivityNet Entities}
\textbf{ActivityNet Entities.} The task in the recently released ActivityNet Entities~\cite{Zhou:CVPR19} dataset, containing 15k videos and more than 158k annotated bounding boxes, is to generate a sentence describing the event in a ground-truth video segment, and to spatially localize all the generated nouns that belong to a vocabulary of $432$ object classes. Following Zhou et al.~\cite{Zhou:CVPR19}, the quality of generated captions is measured using standard metrics, such as Bleu (B@1, B@4), METEOR (M), CIDEr (C), and SPICE (S), whereas the quality of object localization is evaluated on generated sentences using the $F1_{all}$, $F1_{loc}$  metrics. Object localization results on the test set were obtained using the evaluation server~\footnote{\url{https://competitions.codalab.org/competitions/20537}}.
\label{subsec:activitynet}
\begin{table}[t]
\caption{\textbf{Grounded video description results on ActivityNet Entities~\cite{Zhou:CVPR19}}. MHA: multi-head self-attention. SCM-VG: our semantic context module with visual-to-symbolic node correspondences pre-trained on Visual Genome.}
\centering
\scalebox{1}{
\begin{tabular}{l|ccccc|ccc}
\hline
  & $B@1$ & $B@4$ & $M$ & $C$ & $S$ & $F1_{all}$ & $F1_{loc}$\\
  \hline
 \textbf{Validation set}  &  & & &  & &  &  \\ \hline
GVD (MHA)~\cite{Zhou:CVPR19} & \textbf{23.9} & 2.59 & 11.2 & 47.5 & 15.1 & 7.11 & 24.1\\
\hline
GVD (VCM + SCM) (ours) & 23.4 & 2.41 & 11.1 & 47.3 & 14.8 &  7.28 & 25.2\\
GVD (MHA + SCM) (ours) & 23.8 & 2.67 & \textbf{11.3} & 48.6 & \textbf{15.2} & \textbf{7.35} & \textbf{25.3} \\
GVD (MHA + SCM-VG) (ours) & \textbf{23.9}  & \textbf{2.78} & \textbf{11.3} & \textbf{49.1} & 15.1 & 7.15 & 24.0  \\
\hline
\textbf{Test set}  &  & & &  & &  &  & \\ \hline
Masked Transformer~\cite{Zhou:CVPR18} & 22.9 & 2.41 & 10.6 & 46.1 &13.7 & - & - \\
Bi-LSTM+TempoAttn~\cite{Zhou:CVPR18} & 22.8 & 2.17 & 10.2 & 42.2 & 11.8   & - & - \\
GVD (MHA)~\cite{Zhou:CVPR19}  & 23.6 & 2.35 & 11.0 & 45.5 & 14.9  & 7.59 & 25.0\\
\hline
GVD (VCM + SCM) (ours)  & 23.1 & 2.34 & 10.9 & 46.1 & 14.5 & - & -\\
GVD (MHA + SCM) (ours)& 23.6  & 2.54 & 11.2 & 47.7 & 15.0 & 7.30 & 24.4\\
GVD (MHA + SCM-VG) (ours) & \textbf{24.1} & \textbf{2.63} & \textbf{11.4} & \textbf{49.0} & \textbf{15.1} & \textbf{7.81} & \textbf{27.1}\\
\hline
\end{tabular}
}
\label{tab:activitynet}
\end{table}
The current state-of-the-art grounded video description model (GVD)~\cite{Zhou:CVPR19} uses a hierarchical LSTM decoder that generates a sentence describing a video segment, given global video features as well as local region features of $100$ region proposals from $10$ equidistant frames. The region features are refined using a \emph{multi-head self-attention} (MHA) mechanism. To validate the effectiveness of our model, we experiment with three variants of the GVD: (a) replace the MHA with our \ac{STMPNN}; (b) use the MHA along with our Semantic Context Module; (c) the same as before but with visual-to-symbolic node assignment weights initialized based on knowledge transfer from the Visual Genome dataset~\cite{Krishna:IJCV17}. We use a symbolic graph whose nodes correspond to object classes. As shown in Table~\ref{tab:activitynet}, replacing MHA with our visual module does not improve captioning, but it improves localization accuracy with a relative improvement of $4\%$ (24.1 $\rightarrow$ 25.2). Adding our Semantic Context Module to GVD leads to an improvement across all captioning and localization metrics, which is even more pronounced in the test set (improving CIDEr from 45.5 to 47.7). Note that the initial region features already captured semantic information by including object class probabilities. Therefore, the improvement in captioning cannot be attributed solely to the inclusion of semantic context, but rather to our semantic reasoning framework. Finally, from the superior captioning performance of our third variant, we conclude that prior knowledge about correspondences between visual and symbolic nodes, if available, can possibly facilitate representation learning on the hybrid graph.
\section{Conclusions}
\label{sec:conclusions}
In this paper, we proposed a novel deep learning framework for video understanding that performs joint representation learning on a hybrid graph, composed of a symbolic graph and a visual st-graph, for obtaining context-aware visual node and edge features.  We obtained state-of-the-art performance on three challenging datasets, demonstrating the effectiveness and generality of our framework.


\smallskip\noindent\textbf{Acknowledgements:}
The authors thank Carolina Pacheco O\~nate, Paris Giampouras 
and the anonymous reviewers for their valuable comments. This research was 
supported by the IARPA DIVA program via contract number D17PC00345.

%
%
\bibliographystyle{splncs04}
\bibliography{eccv20_bib}

\begin{thebibliography}{10}
\providecommand{\url}[1]{\texttt{#1}}
\providecommand{\urlprefix}{URL }
\providecommand{\doi}[1]{https://doi.org/#1}

\bibitem{Assari:CVPR14}
Assari, S.M., Zamir, A.R., Shah, M.: Video classification using semantic
  concept co-occurrences. In: {IEEE} Conference on Computer Vision and Pattern
  Recognition (2014)

\bibitem{Bajaj:ICCV19}
Bajaj, M., Wang, L., Sigal, L.: G3raphground: Graph-based language grounding.
  In: {IEEE} International Conference on Computer Vision. pp. 4281--4290 (2019)

\bibitem{Baradel:ECCV18}
Baradel, F., Neverova, N., Wolf, C., Mille, J., Mori, G.: Object level visual
  reasoning in videos. In: European Conference on Computer Vision. pp. 106--122
  (2018)

\bibitem{Carreira:CVPR17}
Carreira, J., Zisserman, A.: Quo {Vadis}, {Action} {Recognition}? {A} {New}
  {Model} and the {Kinetics} {Dataset}. In: {IEEE} Conference on Computer
  Vision and Pattern Recognition. pp. 4724--4733 (2017).
  \doi{10.1109/CVPR.2017.502}

\bibitem{Chen:CVPR18}
Chen, X., Li, L., Fei-Fei, L., Gupta, A.: Iterative {Visual} {Reasoning}
  {Beyond} {Convolutions}. In: {IEEE} Conference on Computer Vision and Pattern
  Recognition. pp. 7239--7248 (2018). \doi{10.1109/CVPR.2018.00756}

\bibitem{Chen:CVPR19}
Chen, Y., Rohrbach, M., Yan, Z., Shuicheng, Y., Feng, J., Kalantidis, Y.:
  Graph-based global reasoning networks. In: {IEEE} Conference on Computer
  Vision and Pattern Recognition (2019)

\bibitem{Cheron:ICCV15}
Ch{\'e}ron, G., Laptev, I., Schmid, C.: P-{{CNN}}: {{Pose}}-{{Based CNN
  Features}} for {{Action Recognition}}. In: {IEEE} International Conference on
  Computer Vision. pp. 3218--3226 (2015). \doi{10.1109/ICCV.2015.368}

\bibitem{Choi:CVPR10}
Choi, M.J., Lim, J.J., Torralba, A., Willsky, A.S.: Exploiting hierarchical
  context on a large database of object categories. In: {IEEE} Conference on
  Computer Vision and Pattern Recognition. pp. 129--136 (2010).
  \doi{10.1109/CVPR.2010.5540221}

\bibitem{Dave:CVPR17}
Dave, A., Russakovsky, O., Ramanan, D.: Predictive-{Corrective} {Networks} for
  {Action} {Detection}. In: {IEEE} Conference on Computer Vision and Pattern
  Recognition. pp. 2067--2076 (2017). \doi{10.1109/CVPR.2017.223}

\bibitem{Deng:ECCV14}
Deng, J., Ding, N., Jia, Y., Frome, A., Murphy, K., Bengio, S., Li, Y., Neven,
  H., Adam, H.: Large-{Scale} {Object} {Classification} {Using} {Label}
  {Relation} {Graphs}. In: European Conference on Computer Vision. pp. 48--64.
  Lecture {Notes} in {Computer} {Science} (2014)

\bibitem{Deng:CVPR16}
Deng, Z., Vahdat, A., Hu, H., Mori, G.: Structure {Inference} {Machines}:
  {Recurrent} {Neural} {Networks} for {Analyzing} {Relations} in {Group}
  {Activity} {Recognition}. In: {IEEE} Conference on Computer Vision and
  Pattern Recognition. pp. 4772--4781 (2016). \doi{10.1109/CVPR.2016.516}

\bibitem{Ghosh:WACV20}
Ghosh, P., Yao, Y., Davis, L., Divakaran, A.: Stacked spatio-temporal graph
  convolutional networks for action segmentation. In: IEEE Winter Applications
  of Computer Vision Conference (2020)

\bibitem{Gilmer:ICML17}
Gilmer, J., Schoenholz, S.S., Riley, P.F., Vinyals, O., Dahl, G.E.: Neural
  {Message} {Passing} for {Quantum} {Chemistry}. In: International Conference
  on Machine learning. pp. 1263--1272 (2017)

\bibitem{Girdhar:CVPR19}
Girdhar, R., Carreira, J., Doersch, C., Zisserman, A.: Video action transformer
  network. In: {IEEE} Conference on Computer Vision and Pattern Recognition
  (2019)

\bibitem{Gkioxari:ICCV15}
Gkioxari, G., Girshick, R., Malik, J.: Contextual {Action} {Recognition} with
  {R}*{CNN}. In: {IEEE} International Conference on Computer Vision. pp.
  1080--1088 (2015). \doi{10.1109/ICCV.2015.129}

\bibitem{Gong:CVPR19}
Gong, L., Cheng, Q.: Exploiting edge features for graph neural networks. In:
  {IEEE} Conference on Computer Vision and Pattern Recognition (2019)

\bibitem{He:TPAMI18}
He, K., Gkioxari, G., Dollar, P., Girshick, R.: Mask {R}-{CNN}. {IEEE}
  Transactions on Pattern Analysis and Machine Intelligence pp.~1--1 (2018).
  \doi{10.1109/TPAMI.2018.2844175}

\bibitem{He:ICCV15}
He, K., Zhang, X., Ren, S., Sun, J.: Delving deep into rectifiers: Surpassing
  human-level performance on imagenet classification. In: {IEEE} International
  Conference on Computer Vision (2015)

\bibitem{Huang:BMVC19}
Huang, H., Zhou, L., Zhang, W., Xu, C.: Dynamic {Graph} {Modules} for
  {Modeling} {Higher}-{Order} {Interactions} in {Activity} {Recognition}. In:
  British Machine Vision Conference (2019)

\bibitem{Ibrahim:ECCV18}
Ibrahim, M.S., Mori, G.: Hierarchical relational networks for group activity
  recognition and retrieval. In: European Conference on Computer Vision. pp.
  721--736 (2018)

\bibitem{Jain:CVPR16}
Jain, A., Zamir, A.R., Savarese, S., Saxena, A.: Structural-{RNN}: {Deep}
  {Learning} on {Spatio}-{Temporal} {Graphs}. In: {IEEE} Conference on Computer
  Vision and Pattern Recognition. pp. 5308--5317 (2016)

\bibitem{Jiang:NIPS18}
Jiang, C., Xu, H., Liang, X., Lin, L.: Hybrid {Knowledge} {Routed} {Modules}
  for {Large}-scale {Object} {Detection}. In: Neural Information Processing
  Systems, pp. 1552--1563 (2018)

\bibitem{Jiang:TPAMI18}
Jiang, Y.G., Wu, Z., Wang, J., Xue, X., Chang, S.F.: Exploiting feature and
  class relationships in video categorization with regularized deep neural
  networks. {IEEE} Transactions on Pattern Analysis and Machine Intelligence
  \textbf{40}(2),  352--364 (2018). \doi{10.1109/TPAMI.2017.2670560}

\bibitem{Junior:TPAMI19}
Junior, N.I.N., Hu, H., Zhou, G., Deng, Z., Liao, Z., Mori, G.: Structured
  {Label} {Inference} for {Visual} {Understanding}. {IEEE} Transactions on
  Pattern Analysis and Machine Intelligence pp.~1--1 (2019).
  \doi{10.1109/TPAMI.2019.2893215}

\bibitem{Kipf:ICLR17}
Kipf, T.N., Welling, M.: Semi-supervised classification with graph
  convolutional networks. In: International Conference on Learning
  Representations (2017)

\bibitem{Koller:CVPR94}
Koller, D., Weber, J., Huang, T., Malik, J., Ogasawara, G., Rao, B., Russell,
  S.: Towards robust automatic traffic scene analysis in real-time. In: {IEEE}
  Conference on Computer Vision and Pattern Recognition (1994)

\bibitem{Koppula:IJRR13}
Koppula, H.S., Gupta, R., Saxena, A.: Learning human activities and object
  affordances from rgb-d videos. International Journal Robotics Research
  \textbf{32}(8),  951--970 (2013). \doi{10.1177/0278364913478446}

\bibitem{Krishna:IJCV17}
Krishna, R., Zhu, Y., Groth, O., Johnson, J., Hata, K., Kravitz, J., Chen, S.,
  Kalantidis, Y., Li, L.J., Shamma, D.A., Bernstein, M.S., Fei-Fei, L.: Visual
  genome: Connecting language and vision using crowdsourced dense image
  annotations. International Journal of Computer Vision  \textbf{123}(1),
  32–73 (2017). \doi{10.1007/s11263-016-0981-7}

\bibitem{Lea:CVPR17}
Lea, C., Flynn, M.D., Vidal, R., Reiter, A., Hager, G.D.: Temporal
  {Convolutional} {Networks} for {Action} {Segmentation} and {Detection}. In:
  {IEEE} Conference on Computer Vision and Pattern Recognition. pp. 1003--1012
  (2017). \doi{10.1109/CVPR.2017.113}

\bibitem{Lee:CVPR18}
Lee, C., Fang, W., Yeh, C., Wang, Y.F.: Multi-label {Zero}-{Shot} {Learning}
  with {Structured} {Knowledge} {Graphs}. In: {IEEE} Conference on Computer
  Vision and Pattern Recognition. pp. 1576--1585 (2018).
  \doi{10.1109/CVPR.2018.00170}

\bibitem{Li:ICCV19}
Li, K., Zhang, Y., Li, K., Li, Y., Fu, Y.: Visual semantic reasoning for
  image-text matching. In: {IEEE} International Conference on Computer Vision
  (2019)

\bibitem{Li:ICCV17}
Li, R., Tapaswi, M., Liao, R., Jia, J., Urtasun, R., Fidler, S.: Situation
  recognition with graph neural networks. In: {IEEE} International Conference
  on Computer Vision (2017)

\bibitem{Li:NIPS18}
Li, Y., Gupta, A.: Beyond grids: Learning graph representations for visual
  recognition. In: Bengio, S., Wallach, H., Larochelle, H., Grauman, K.,
  Cesa-Bianchi, N., Garnett, R. (eds.) Neural Information Processing Systems.
  pp. 9225--9235 (2018)

\bibitem{Liang:NIPS18}
Liang, X., Hu, Z., Zhang, H., Lin, L., Xing, E.P.: Symbolic {Graph} {Reasoning}
  {Meets} {Convolutions}. In: Neural Information Processing Systems. pp.
  1853--1863. Curran Associates, Inc. (2018)

\bibitem{Lin:ECCV14}
Lin, T.Y., Maire, M., Belongie, S., Hays, J., Perona, P., Ramanan, D.,
  Doll{\'a}r, P., Zitnick, C.L.: Microsoft coco: Common objects in context. In:
  European Conference on Computer Vision. pp. 740--755. Springer International
  Publishing, Cham (2014)

\bibitem{Liu:CVPR11}
Liu, J., Kuipers, B., Savarese, S.: Recognizing human actions by attributes.
  In: {IEEE} Conference on Computer Vision and Pattern Recognition. pp.
  3337--3344 (2011). \doi{10.1109/CVPR.2011.5995353}

\bibitem{Ma:CVPR18}
Ma, C., Kadav, A., Melvin, I., Kira, Z., AlRegib, G., Graf, H.P.: Attend and
  {Interact}: {Higher}-{Order} {Object} {Interactions} for {Video}
  {Understanding}. In: {IEEE} Conference on Computer Vision and Pattern
  Recognition. pp. 6790--6800 (2018). \doi{10.1109/CVPR.2018.00710}

\bibitem{Marszalek:CVPR07}
Marszalek, M., Schmid, C.: Semantic {Hierarchies} for {Visual} {Object}
  {Recognition}. In: {IEEE} Conference on Computer Vision and Pattern
  Recognition. pp.~1--7 (2007). \doi{10.1109/CVPR.2007.383272}

\bibitem{Marszalek:ECCV08}
Marszalek, M., Schmid, C.: Constructing {Category} {Hierarchies} for {Visual}
  {Recognition}. In: European Conference on Computer Vision. pp. 479--491.
  Springer-Verlag, Berlin, Heidelberg (2008)

\bibitem{Mavroudi:WACV17}
Mavroudi, E., Tao, L., Vidal, R.: Deep {Moving} {Poselets} for {Video} {Based}
  {Action} {Recognition}. In: IEEE Winter Applications of Computer Vision
  Conference. pp. 111--120 (2017). \doi{10.1109/WACV.2017.20}

\bibitem{Mikolov:NIPS13}
Mikolov, T., Sutskever, I., Chen, K., Corrado, G.S., Dean, J.: Distributed
  {Representations} of {Words} and {Phrases} and their {Compositionality}. In:
  Neural Information Processing Systems, pp. 3111--3119 (2013)

\bibitem{Nicolicioiu:NIPS19}
Nicolicioiu, A., Duta, I., Leordeanu, M.: Recurrent space-time graph neural
  networks. In: Neural Information Processing Systems (2019)

\bibitem{Oliva:TCS07}
Oliva, A., Torralba, A.: The role of context in object recognition. Trends in
  Cognitive Sciences  \textbf{11}(12),  520 -- 527 (2007).
  \doi{https://doi.org/10.1016/j.tics.2007.09.009}

\bibitem{Piergiovanni:CVPR18}
Piergiovanni, A., Ryoo, M.S.: Learning {Latent} {Super}-{Events} to {Detect}
  {Multiple} {Activities} in {Videos}. In: {IEEE} Conference on Computer Vision
  and Pattern Recognition. pp. 5304--5313 (2018). \doi{10.1109/CVPR.2018.00556}

\bibitem{Piergiovanni:ICML19}
Piergiovanni, A.J., Ryoo, M.S.: Temporal gaussian mixture layer for videos. In:
  International Conference on Machine learning (2019)

\bibitem{Prest:TPAMI12}
{Prest}, A., {Ferrari}, V., {Schmid}, C.: Explicit modeling of human-object
  interactions in realistic videos. {IEEE} Transactions on Pattern Analysis and
  Machine Intelligence  (2013)

\bibitem{Qi:ECCV18}
Qi, S., Wang, W., Jia, B., Shen, J., Zhu, S.C.: Learning human-object
  interactions by graph parsing neural networks. In: European Conference on
  Computer Vision. pp. 401--417 (2018)

\bibitem{Ramanathan:CVPR15}
Ramanathan, V., Li, C., Deng, J., Han, W., Li, Z., Gu, K., Song, Y., Bengio,
  S., Rossenberg, C., Fei-Fei, L.: Learning semantic relationships for better
  action retrieval in images. In: {IEEE} Conference on Computer Vision and
  Pattern Recognition. pp. 1100--1109 (2015). \doi{10.1109/CVPR.2015.7298713}

\bibitem{Schlichtkrull:ESWC18}
Schlichtkrull, M., Kipf, T.N., Bloem, P., Van Den~Berg, R., Titov, I., Welling,
  M.: Modeling relational data with graph convolutional networks. In: European
  Semantic Web Conference. pp. 593--607. Springer (2018)

\bibitem{Sigurdsson:CVPR17}
Sigurdsson, G.A., Divvala, S., Farhadi, A., Gupta, A.: Asynchronous {Temporal}
  {Fields} for {Action} {Recognition}. In: {IEEE} Conference on Computer Vision
  and Pattern Recognition. pp. 5650--5659 (2017). \doi{10.1109/CVPR.2017.599}

\bibitem{Sigurdsson:ECCV16}
Sigurdsson, G.A., Varol, G., Wang, X., Farhadi, A., Laptev, I., Gupta, A.:
  Hollywood in homes: Crowdsourcing data collection for activity understanding.
  In: European Conference on Computer Vision. pp. 510--526 (2016)

\bibitem{Simonyan:NIPS14}
Simonyan, K., Zisserman, A.: Two-stream convolutional networks for action
  recognition in videos. In: Ghahramani, Z., Welling, M., Cortes, C., Lawrence,
  N.D., Weinberger, K.Q. (eds.) Neural Information Processing Systems. pp.
  568--576. Curran Associates, Inc. (2014)

\bibitem{Sun:ECCV18}
Sun, C., Shrivastava, A., Vondrick, C., Murphy, K., Sukthankar, R., Schmid, C.:
  Actor-centric relation network. In: European Conference on Computer Vision.
  pp. 318--334 (2018)

\bibitem{Teney:CVPR17}
Teney, D., Liu, L., van~den Hengel, A.: Graph-structured representations for
  visual question answering. In: {IEEE} Conference on Computer Vision and
  Pattern Recognition. pp.~1--9 (2017)

\bibitem{Tran:ICCV15}
Tran, D., Bourdev, L., Fergus, R., Torresani, L., Paluri, M.: Learning
  spatiotemporal features with 3d convolutional networks. In: {IEEE}
  International Conference on Computer Vision (2015)

\bibitem{Velickovic:ICLR18}
Veli{\v{c}}kovi{\'{c}}, P., Cucurull, G., Casanova, A., Romero, A., Li{\`{o}},
  P., Bengio, Y.: {Graph Attention Networks}. International Conference on
  Learning Representations  (2018)

\bibitem{Wang:ECCV16}
Wang, L., Xiong, Y., Wang, Z., Qiao, Y., Lin, D., Tang, X., {Val Gool}, L.:
  Temporal segment networks: Towards good practices for deep action
  recognition. In: European Conference on Computer Vision (2016)

\bibitem{Wang:ECCV18}
Wang, X., Gupta, A.: Videos as space-time region graphs. In: European
  Conference on Computer Vision. pp. 413--431 (2018)

\bibitem{Wang:CVPR15}
Wang, X., Ji, Q.: Video event recognition with deep hierarchical context model.
  In: {IEEE} Conference on Computer Vision and Pattern Recognition (2015)

\bibitem{Xiong:ICCV19}
Xiong, Y., Huang, Q., Guo, L., Zhou, H., Zhou, B., Lin, D.: A graph-based
  framework to bridge movies and synopses. In: {IEEE} International Conference
  on Computer Vision (2019)

\bibitem{Xu:ICCV17}
Xu, H., Das, A., Saenko, K.: R-{C}3d: {Region} {Convolutional} 3d {Network} for
  {Temporal} {Activity} {Detection}. In: {IEEE} International Conference on
  Computer Vision. pp. 5794--5803 (2017). \doi{10.1109/ICCV.2017.617}

\bibitem{Yatskar:CVPR16}
Yatskar, M., Zettlemoyer, L., Farhadi, A.: Situation recognition: Visual
  semantic role labeling for image understanding. In: {IEEE} Conference on
  Computer Vision and Pattern Recognition (2016)

\bibitem{Yu:NIPS19}
Yu, W., Zhou, J., Yu, W., Liang, X., Xiao, N.: Heterogeneous graph learning for
  visual commonsense reasoning. In: Advances in Neural Information Processing
  Systems 32. pp. 2769--2779. Curran Associates, Inc. (2019)

\bibitem{Yuan:ICCV17}
Yuan, Y., Liang, X., Wang, X., Yeung, D., Gupta, A.: Temporal {Dynamic} {Graph}
  {LSTM} for {Action}-{Driven} {Video} {Object} {Detection}. In: {IEEE}
  International Conference on Computer Vision. pp. 1819--1828 (2017).
  \doi{10.1109/ICCV.2017.200}

\bibitem{Zhang:CVPR19}
Zhang, Y., Tokmakov, P., Hebert, M., Schmid, C.: A structured model for action
  detection. In: {IEEE} Conference on Computer Vision and Pattern Recognition
  (2019)

\bibitem{Zhou:ECCV18}
Zhou, B., Andonian, A., Oliva, A., Torralba, A.: Temporal relational reasoning
  in videos. In: Computer Vision -- ECCV 2018. pp. 831--846 (2018)

\bibitem{Zhou:CVPR19}
Zhou, L., Kalantidis, Y., Chen, X., Corso, J.J., Rohrbach, M.: Grounded video
  description. In: {IEEE} Conference on Computer Vision and Pattern Recognition
  (2019)

\bibitem{Zhou:CVPR18}
Zhou, L., Zhou, Y., Corso, J.J., Socher, R., Xiong, C.: End-to-end dense video
  captioning with masked transformer. In: {IEEE} Conference on Computer Vision
  and Pattern Recognition. pp. 8739--8748 (2018)

\bibitem{Zhou:CVPR15}
Zhou, Y., Ni, B., {and}, Tian, Q.: Interaction part mining: {A} mid-level
  approach for fine-grained action recognition. In: {IEEE} Conference on
  Computer Vision and Pattern Recognition. pp. 3323--3331 (2015).
  \doi{10.1109/CVPR.2015.7298953}

\bibitem{Zhu:CVPR13}
Zhu, Y., Nayak, N.M., Roy-Chowdhury, A.K.: Context-aware modeling and
  recognition of activities in video. In: {IEEE} Conference on Computer Vision
  and Pattern Recognition (2013)

\bibitem{Zitnik:Bioinformatics18}
Zitnik, M., Agrawal, M., Leskovec, J.: Modeling polypharmacy side effects with
  graph convolutional networks. Bioinformatics p. 457–466 (2018)

\end{thebibliography}
\end{document}